\documentclass[letterpaper, 10 pt, conference]{ieeeconf}  % Comment this line out if you need a4paper

\IEEEoverridecommandlockouts                              % This command is only needed if 
\title{\LARGE \bf
Enhancing Safety for Autonomous Agents in Partly Concealed Urban Traffic Environments Through Representation-Based Shielding
}
\author{Pierre Haritz$^{1,*}$, David Wanke$^{1,*}$ and Thomas Liebig$^{1,2,*}$% <-this % stops a space
\thanks{$^{1}$Faculty of Computer Science, Chair of Artificial Intelligence, TU Dortmund University, Dortmund, Germany}%
\thanks{$^{2}$Lamarr Institute for Machine Learning and Artificial Intelligence, Dortmund, Germany}
\thanks{$^{*}$ {\tt\small firstname.lastname@tu-dortmund.de}}
\thanks{This work was supported by the Federal Ministry of Education and Research of Germany and the state of North-Rhine Westphalia as part of the Lamarr-Institute for Machine Learning and Artificial Intelligence.}% <-this % stops a space
\thanks{\textit{©2024 IEEE.  Personal use of this material is permitted.  Permission from IEEE must be obtained for all other uses, in any current or future media, including reprinting/republishing this material for advertising or promotional purposes, creating new collective works, for resale or redistribution to servers or lists, or reuse of any copyrighted component of this work in other works.}}
}

\usepackage{cite}
\usepackage{amsmath,amssymb,amsfonts}
\usepackage{algorithmic}
\usepackage{graphicx}
\usepackage{textcomp}
\usepackage{xcolor}
\usepackage{flushend}
\usepackage{bbm}
\usepackage{booktabs}

\begin{document}

\maketitle
\thispagestyle{empty}
\pagestyle{empty}

%%%%%%%%%%%%%%%%%%%%%%%%%%%%%%%%%%%%%%%%%%%%%%%%%%%%%%%%%%%%%%%%%%%%%%%%%%%%%%%%

\begin{abstract}
Navigating unsignalized intersections in urban environments poses a complex challenge for self-driving vehicles, where issues such as view obstructions, unpredictable pedestrian crossings, and diverse traffic participants demand a great focus on crash prevention. In this paper, we propose a novel state representation for Reinforcement Learning (RL) agents centered around the information perceivable by an autonomous agent, enabling the safe navigation of previously uncharted road maps.

Our approach surpasses several baseline models by a significant margin in terms of safety and energy consumption metrics. These improvements are achieved while maintaining a competitive average travel speed. Our findings pave the way for more robust and reliable autonomous navigation strategies, promising safer and more efficient urban traffic environments.
%Finding a strategy for self-driving vehicles to navigate unsignaled intersections in urban environments can be a difficult task when simultaneously dealing with view obstructions, the possibility of crossing pedestrians and other traffic participants, that makes the prevention of crashes a primary concern. We propose a state representation for Reinforcement Learning agents based on information perceivable by our agent that makes it possible to safely navigate previously unseen road maps. Our approach manages to outperform several baselines significantly on safety and energy consumption metrics, while achieving similar average travel speed.
\end{abstract}

%\begin{IEEEkeywords}
%reinforcement learning, safety, control, generalizing, traffic environment, autonomous driving
%\end{IEEEkeywords}

\section{Introduction}
The prominence of autonomous driving in contemporary society has surged significantly. This transformation is propelled by the recognition that the operational environment for autonomous vehicles is characterized by an exceptional degree of complexity and continuous dynamism. Traditional rule-based control systems, which are often predicated on fixed sets of instructions and heuristics, encounter their limitations when confronted with the multifaceted intricacies of real-world driving scenarios. As a result, there is a compelling imperative for the integration of machine learning (ML) approaches to facilitate the adaptive, intelligent decision-making required for navigating this intricate landscape effectively.

Machine learning equips autonomous vehicles with the capacity to learn and adapt from data, thereby enabling them to respond to the ever-evolving challenges of the road. By harnessing ML, these vehicles can harness the power of data-driven insights to make nuanced, real-time decisions. These decisions encompass a wide spectrum of scenarios, from navigating through dense urban traffic and responding to sudden road hazards to anticipating the behavior of other road users, including pedestrians and fellow motorists. The integration of ML techniques has ushered in a new era of autonomous driving, where vehicles can actively learn from their experiences and continually improve their decision-making prowess, ultimately enhancing the safety, efficiency, and reliability of autonomous transportation systems.
%Autonomous driving increasingly plays a bigger role in today's society. Since the environment in which an autonomous vehicle is required to operate is highly complex and dynamic, rule-based control is at its capacity, and therefore, machine learning (ML) approaches are necessary.

Reinforcement Learning (RL) represents a machine learning paradigm tailored for addressing sequential decision-making challenges, a domain closely resembling the decision-making tasks encountered by autonomous vehicles. Within this framework, an RL agent actively interacts with its environment, gradually discerning the value associated with its actions, and ultimately devising an effective control strategy. In real-world scenarios like autonomous driving, safety assumes paramount importance.

In the context of our study, we focus on the intricate scenario of navigating urban environments. Here, self-driving cars, referred to as 'agents,' must navigate while adhering to the rules of the road and respecting the presence of other traffic participants, encompassing both vehicles and pedestrians. The utmost priority in this context is the avoidance of collisions, even during the initial exploratory phase when the agent is actively learning an optimal decision-making policy. This includes navigating situations where certain elements of the relevant environment may be temporarily obscured, such as by obstructive infrastructure or visual obstacles. Ensuring safety during these circumstances represents a fundamental challenge within the realm of autonomous driving.
%Reinforcement Learning (RL) is a paradigm of ML that deals with sequential decision-making tasks like those an autonomous vehicle must make. By interacting with an environment, an RL agent can learn the value that their actions hold and is thereby able to devise a strategy for control in such. Importantly, especially in real-world scenarios such as autonomous driving, safety plays a huge, and arguably the most important, role. We consider a scenario of driving in an urban environment, where self-driving cars (\textit{agents}) are subject to respecting other traffic participants, be they cars or pedestrians. Naturally, any collisions must be prevented even during an exploratory phase, where the agent is still trying to learn an optimal policy. This includes situations where only part of the relevant environment can be observed, e.g., by occluding infrastructure.
In this research, we introduce a novel safety-oriented shielding strategy tailored specifically for \textit{Deep~Q-Learning}. Our approach is designed to excel in a scenario where observations are comprehended via an invariant representation of the environment. Our study offers several noteworthy contributions:
\begin{itemize}
    \item \textbf{Safety-Centric Shielding:}  We present an innovative safety-centric shielding technique, forming the paper-at-hands core.
    \item \textbf{Generalization and Adaptability:} Our findings underscore the agent's ability to generalize its learned policies effectively. This critical aspect showcases the agent's capability to navigate previously unseen road maps, addressing the challenge of view obstructions, pedestrian interactions, and diverse traffic scenarios.
    \item \textbf{Collision Mitigation:} One of the primary objectives emphasized above is the prevention of collisions, even during the exploration phase of the reinforcement learning algorithm. Our research demonstrates the agent's ability to significantly reduce collisions compared to less safe baseline approaches, underscoring its effectiveness in enhancing safety metrics.
    \item \textbf{Efficiency and Speed:} While prioritizing safety, our approach maintains a comparable average travel speed to less safe baselines. This achievement aligns with our focus on achieving both safety and energy consumption metrics without sacrificing travel efficiency.
\end{itemize}
Our research introduces a novel shielding strategy for Deep Q-Learning, showcasing its efficacy in enhancing safety, generalization, collision mitigation, and maintaining travel efficiency. These contributions collectively advance the state-of-the-art in autonomous intersection navigation.

The paper is structured as follows. Section 2 introduces preliminaries: Reinforcement Learning and Q-Learning and Model Checking and Safety in Reinforcement Learning. Related Work is in Section 3. Our methodology is presented afterward in Section 4, followed by an evaluation in Section 5. We conclude in the last section.
%In this paper, we propose a safe shielding approach for \textit{Deep~Q~Learning} and for a scenario in which observations get perceived through an invariant environment representation. We show that our agent generalizes well and can navigate through the street system of randomly generated environments with minimal collisions while achieving a similar average speed compared to unsafe baselines.

\section{Preliminaries}
This section provides the necessary preliminaries to follow the rest of our paper. We Introduce Reinforcement Learning and Deep Q-Learning as a specific algorithm. Afterward, Model Checking and Safety via Shielding for Reinforcement Learning is introduced.
\subsection{Reinforcement Learning}
Reinforcement Learning problems~\cite{sutton2018rl} can typically be modeled as a Markov Decision Process (MDP) $\mathcal{M} = (\mathcal{S}, \mathcal{A}, T, \gamma, R)$ with a state space $\mathcal{S}$, an action space $\mathcal{A}$, a transition probability function $T: \mathcal{S} \times \mathcal{A} \times \mathcal{S} \rightarrow [0,1]$, a discount factor $\gamma \in [0,1]$ and a reward function $R: \mathcal{S} \times \mathcal{A} \rightarrow \mathbb{R}$.

\subsection{Q-Learning}
Q-learning is a fundamental reinforcement learning algorithm designed for agents to learn optimal strategies in sequential decision-making tasks~\cite{sutton2018rl}. At its core is the Q-function, which assigns a value to each state-action pair, representing the expected cumulative reward an agent can attain by taking a specific action from a given state and thereafter following an optimal policy for an MDP. The algorithm aims to learn the optimal action-value function (Q-function) by iteratively updating Q-values based on the observed experiences of an agent interacting with an environment. Q-learning enables agents to make informed decisions by selecting actions that maximize the expected cumulative discounted reward.

The Q-function, denoted as $Q(s, a)$, is a mapping that associates a value with each state-action pair $(s, a)$, where $s$ represents a state in the environment and $a$ represents an action available in that state. The Q-value represents the expected cumulative discounted reward that an agent can achieve by taking action $a$ in state $s$ and thereafter following an optimal policy. Formally, 
\begin{equation}
    Q(s, a) = \mathbb{E}[R_t + \gamma * \max_a Q(s', a) | s, a],
\end{equation}
where $R_t$ is the immediate reward after taking action $a$ in state $s$, $\gamma$ is the discount factor, and $s'$ is the resulting state.

Deep~Q~Networks (DQNs,~\cite{mnih2015human}) are an extension of Q-learning that leverages deep neural networks to approximate the Q-function in high-dimensional state spaces. A DQN consists of a neural network with its inputs representing the state and outputs representing Q-values for each possible action. DQNs use gradient descent to minimize the temporal difference error between the current Q-value estimate and the updated Q-value estimate using the Bellman equation. This allows DQNs to handle complex environments and learn optimal policies directly from raw sensory inputs, such as images.

\subsection{Model Checking}
System models are often represented through transition systems~\cite{keller1976formal}.
A transition system is defined as a tuple \\$(S, Act, \rightarrow, I, AP, L)$, where $S$ is the set of system states, $Act$ is the set of actions in the system, $\rightarrow\subseteq S \times Act \times S$ is the transition relation, $I \subseteq S$ is the set of initial states, $AP$ is the set of atomic propositions and $L: S \rightarrow 2^{AP}$ is a function that maps states to their set of atomic propositions~\cite{baier2008principles}.

A state $s \in S$ fulfills a logical formula $\Phi$ if $L(s)$ also fulfills $\Phi$:
\begin{equation}
    s \models \Phi \Leftrightarrow L(s) \models \Phi
\end{equation}

Such systems can be verified through \textit{model checking} techniques, which is a method for checking whether a finite-state model of a system meets a given specification.

Linear Temporal Logic extends modal logic with additional time modalities~\cite{huth2004logic}.
For a set $e \in AP$, an LTL formula can be defined with a grammar
\begin{equation}
    \varphi ::= true | e | \varphi_1 \lor \varphi_2 | \neg \varphi | \bigcirc \varphi | \cup \varphi.
\end{equation}
A formula $\bigcirc \varphi$ is satisfied in the current time step if the formula $\varphi$ is satisfied in the next time step. On the other hand, a formula $\varphi_1 \cup \varphi_2$ is satisfied in the current time step if there is a future time step in which $\varphi_2$ holds and $\varphi_1$ holds in all time steps up to that future time step.
\subsection{Safety}
Shielding~\cite{alshiekh2018safe} is a concept to prevent critical states for RL agents during both exploration and execution and can be understood as a form of \textit{model checking}, and we can use an LTL formula $\varphi$ to define constraints that the agents need to adhere to for each interaction with the system, or specifically, the environment. It is possible to transfer an MDP $\mathcal{M}$ to a transition system $TS^{\mathcal{M}}$. Analogously, it is possible to model a transition system $TS^\varphi$ from a logical formula $\varphi$. From two transition systems $TS^{\mathcal{M}}$ and $TS^\varphi$, an environment with a shield can be automatically constructed.

\section{Related Work}
Kiran et al. categorize modern autonomous driving systems into scene understanding, decision-making and planning~\cite{kiran2021deep}.
Navigating partly concealed intersections is highlighted in~\cite{isele2018navigating}. The authors show that RL agents are able to cross intersections faster than rule-based vehicles.

Bouton et al.~\cite{bouton2019reinforcement} propose a method that makes it possible to give probabilistic guarantees for the fulfillment of \textit{LTL} formulas for RL agents. Full information about the transition probabilities $T(s'|s,a)$ of the underlying MDP is a requirement, which makes it unfeasible for real-world scenarios. The authors argue, similarly to~\cite{alshiekh2018safe}, that abstraction is sufficient, which we deem a rather hard task for complex scenarios like urban traffic.

In reality, autonomous vehicles perceive their environment through sensors and thus are subject to faulty information such as measurement errors. MDPs, as a process model, assume that the real state is always known. A more viable option to model such problems is to add a belief $b: \mathcal{S} \rightarrow [0,1]$ over the knowledge of environment states, resulting in a process model called \textit{Partially-Observable Markov Decision Process}, or \textit{POMDP}~\cite{kaelbling1998planning}. 
Bouton et al.~\cite{bouton2019safe} include these perception errors in their research and use scene decomposition to enable scaling for situations with multiple road users. In doing so, the complex situation can be decomposed into n different instances of the example scenario.

For large state spaces, for an agent to be able to generalize between states is arguably of value. Kurzer et al.~\cite{kurzer2021generalizing} propose using an invariant representation of the environment for successful intersection navigation.

Traffic rules, such as ``\textit{Left yields to right}'', assign right-of-way priorities to specific traffic participants depending on the situation. Zhang et al.~\cite{zhang2022rule} present an approach for checking such priorities with a rule-base module. For added safety, the Resposibility-Sensitive Safety model~\cite{shalev2017formal} is able to guarantee safety for the decisions of traffic participants. Mokthari and Wagner\cite{mokhtari2021safe} propose a combined rule-based RL approach for navigating unsignalized intersections in fully observable high-fidelity environments.
\section{Methodology}
In this section, we will present our approach of using a safety shield based on a local environment representation.

In order to process the observations that the agent receives, a suitable state representation is needed that encodes all the necessary information for decision-making. The so-called Invariant Environment Representation (IER)~\cite{kurzer2021generalizing} serves as the basis for the state representation.
We propose a new representation based on the agent's ego perspective that holds all available and relevant information for the agent's decision-making, called \textit{IER+}. We first divide the agent's field of vision into sub-spaces of the same size with the visible field's width correlating to the agent vehicle's width. We further encode several types of information into each sub-space: Time-to-Occupancy (TTO) and Time-to-Vacancy (TTV).
Based on the vehicular velocity $v$ and section markers $s_{start}, s_{end}$, we define the time it takes for a vehicle to arrive at a section as
\[
TTO := \frac{s_{start}}{v},
\]
and the time it takes to leave a section as
\[
TTV := \frac{s_{end}}{v}.
\]
Furthermore, $1_{intersection}$ is an indicator function that signals the beginning and the end of an intersection, and $\varphi_{priority, other}$ signals whether the next arriving road user has the right of way, according to the implemented traffic rule set.

\begin{figure}[!h]
    \begin{center}
    \includegraphics[width=1.0\columnwidth]{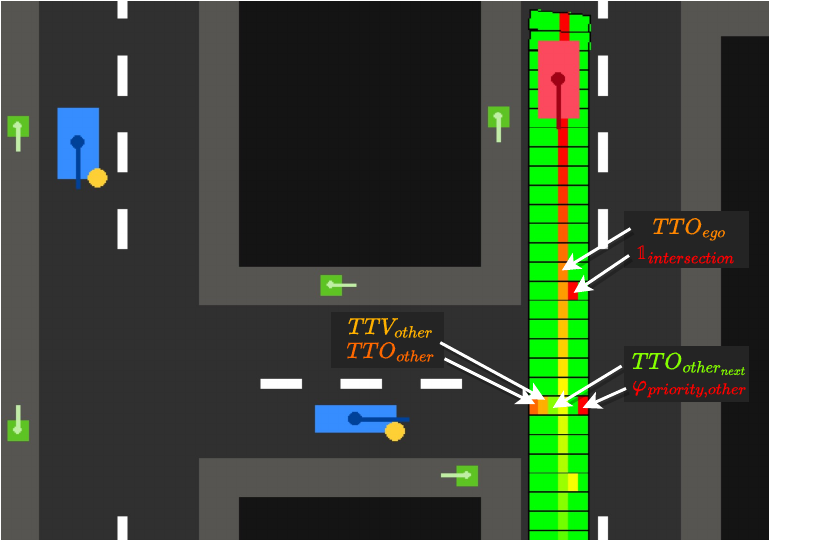}
    %\quelle{Aufnahme aus dem Simulator}
    \caption{Illustration of the state representation with $TTO$ and $TTV$ as colored gradient from red ($t\rightarrow 0$) to green ($t\rightarrow t_{max}$). We color the indicator function $\mathbbm{1}_{intersection}$ as green (neither beginning nor end of the intersection), yellow (intersection end), or red (intersection start). The priority indicator $\varphi_{priority,other}$ is shown as green (the agent has the right of way) or rot (another traffic participant has the right of way).}
    \label{fig:zustandsdarstellung}
    \end{center}
\end{figure}

A visualization of how this information gets encoded into the state representation can be seen in Fig.~\ref{fig:zustandsdarstellung}.

The time it takes for a vehicle $i$ to arrive at an intersection is calculated by $TTI_i=\frac{d_{c,i}}{v_i}$, where $d_{c,i}$ equates to the distance to the \textit{intersection conflict zone} $C$ and $v_i$ is the vehicular velocity. Since the agent has to assume the worst case of a standing vehicle at the intersection that accelerates at a maximum rate. An arriving vehicle $i$ has to fulfill $\varphi_{incoming, i}$:
\[
\varphi_{incoming, i}=\lbrace d_{c,i}\leq D_m \rbrace \wedge \lbrace TTI_i \leq T_a \rbrace
\]
Here, $D_m$ describes the monitoring range of the simulated vehicles that follow the behavioral patterns of a rule-based \textit{IDM}.

\begin{figure}[!htb]
    \begin{center}
    \includegraphics[width=1.0\columnwidth]{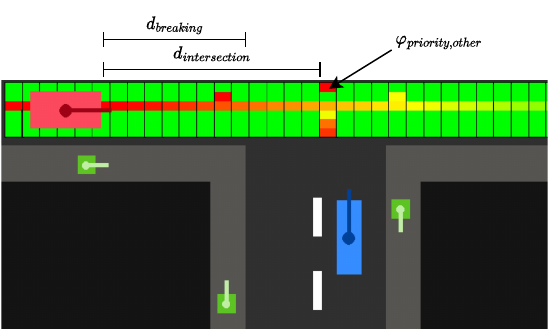}
    %\quelle{Aufnahme aus dem Simulator}
    \caption{Illustration of our \textit{Shielding} approach in \textit{PyAD-RL}. The agent (red) gets supported by a shield that triggers if certain conditions are met.}
    \label{fig:simulation_shielding}
    \end{center}
\end{figure}

We present a \textit{shield} based on \textit{IER+} that both prevents collisions and takes into account the property of minimal interference of shielding.
In this context, Post-Posed Shielding is used, which means that the shield is implemented after the agent has chosen an action. In an emergency, the shield intervenes and overrides the agent's action with the safest action. The advantage of this variant is that it is completely independent of the learning algorithm used~\cite{alshiekh2018safe}. 
For the shield, we define a trigger condition for an \textit{emergency brake} action:
\[
\lbrace 0 \leq d_{intersection} - d_{braking} < d_{threshold} \rbrace \wedge \lbrace \varphi_{priority, other} \rbrace,
\]
where $d_{intersection}$ equates to the distance to the patch, where another traffic participant intersects with the path of the agent, $d_{braking}$ to the stopping distance, and $d_{threshold}$ defines a configurable threshold that includes the safety gap to the preceding vehicle.
Our shielding approach is visualized in figure \ref{fig:simulation_shielding}. 

\begin{figure}[!h]
    \begin{center}
    \includegraphics[width=1.0\columnwidth]{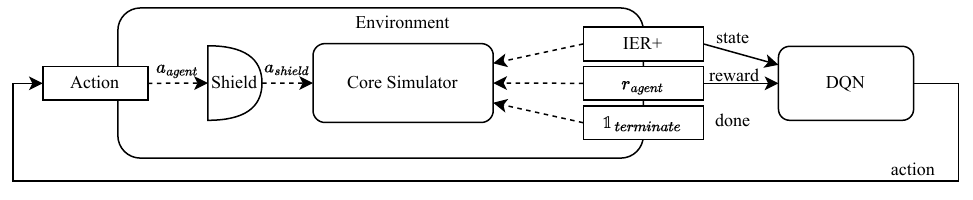}
    %\quelle{Eigene Darstellung}
    \caption{Our IER+-Shielding based RL loop. The selected action gets checked by the shield condition and replaced if necessary. Furthermore, using the shield gets penalized by the reward function to encourge safe control without relying on the shield.}
    \label{fig:rl_zusammensetzung}
    \end{center}
\end{figure}

Since one of our approach's strengths is its modularity, it is decoupled from the choice of the actual RL method, which can be seen in figure \ref{fig:rl_zusammensetzung}. 

\section{Evaluation}
Due to the necessary reproducibility of the action-selection process and the infeasibility of repeating and comparing a process under the same conditions in the real world, Reinforcement Learning problems are typically evaluated in simulated environments, e.g., OpenAI gym~\cite{brockmann2016gym}. In this section, we present our own simulation environment and evaluate our approach in a variety of scenarios. It was specifically developed for RL tasks that connect well to common \textit{python} RL libraries. The code can be found in the \textit{PyAD-RL repository} \footnote{https://github.com/DavidWanke/PyAutoDriveRL}.

\subsection{PyAD-RL Simulation Environment}
Common traffic simulators include SUMO~\cite{wiessner2018sumo} and CARLA~\cite{dosovitskiy2017carla}. While CARLA is a powerful 3D simulator that, among other things, is able to generate realistic images as an environmental representation for reinforcement learning agents, it is highly complex and, therefore, computationally expensive to run. SUMO, on the other hand, is a 2D microscopic traffic simulator. While it is widely used and has its own RL framework (FLOW~\cite{kheterpal2018flow}), functionality such as crossing pedestrians, visibility calculation, and random environment generation are either not supported or not implemented.
\begin{figure}[!h]
    \begin{center}
    \includegraphics[width=1.0\columnwidth]{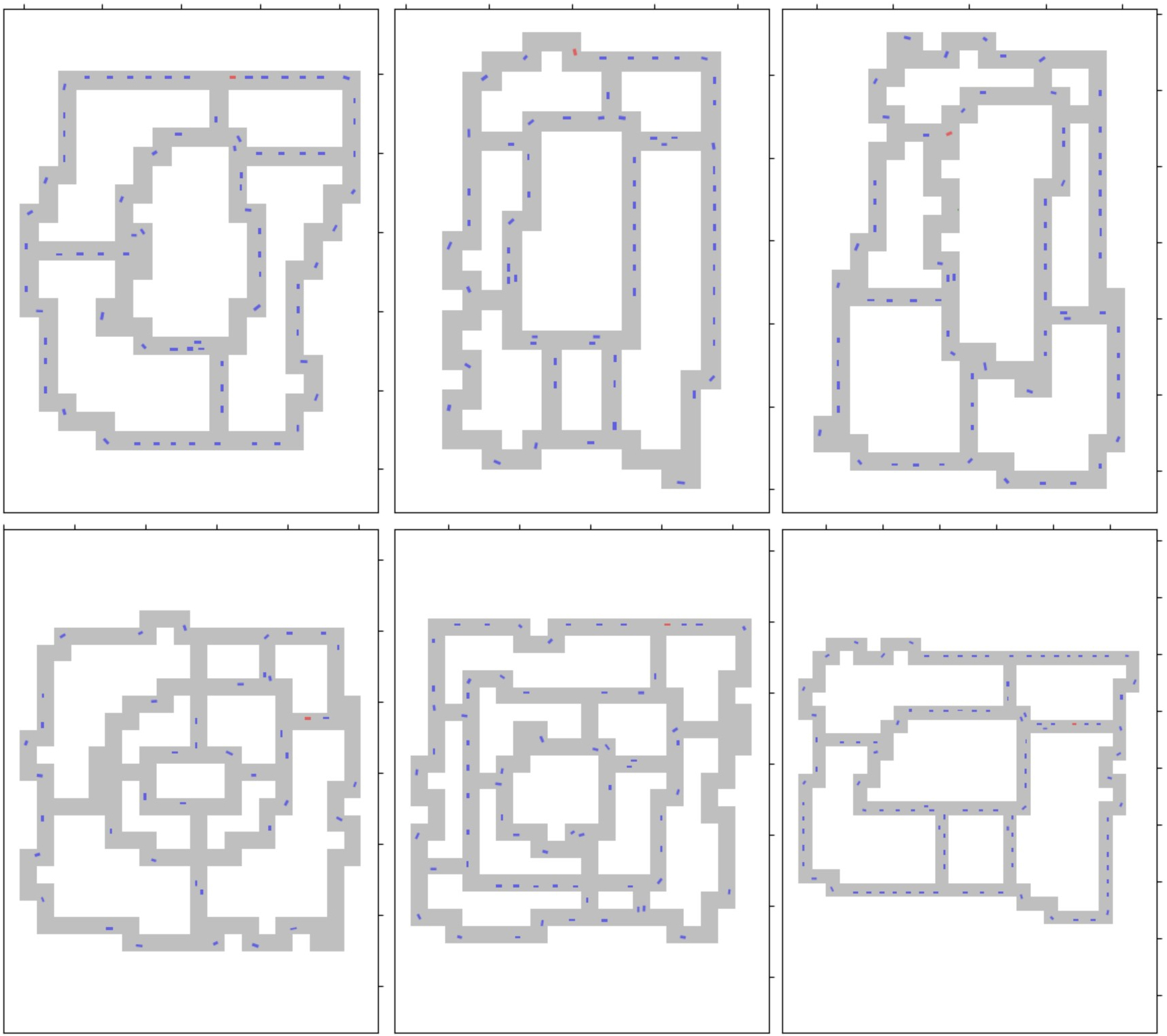}
    %\quelle{Randomly generated urban traffic environments.}
    \caption{Randomly generated urban traffic environments in \textit{PyAD-RL}. Blue vehicles are simulated vehicles with rule-based IDM behavior and the red vehicle is controlled by the agent policy.}
    \label{fig:subprozesse}
    \end{center}
\end{figure}

\subsection{Experiments}
\begin{figure}[!htb]
    \begin{center}
    \includegraphics[width=1.0\columnwidth]{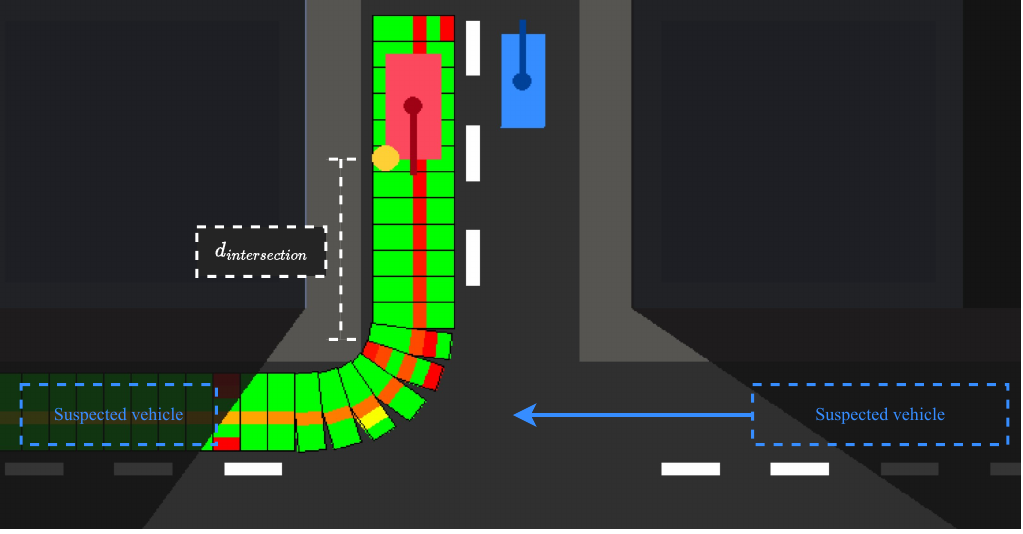}
    %\quelle{Aufnahme aus dem Simulator}
    \caption{Illustration of suspected vehicles at partly concealed intersections in the \textit{PyAD-RL} environment. Here, $d_{intersection}$ also serves as the necessary distance measure for the IDM model.}
    \label{fig:ier-idm-agent}
    \end{center}
\end{figure}

We evaluate our agents in urban traffic environments with possible concealment by buildings, as can be seen in Fig.~\ref{fig:ier-idm-agent}. The agent assumes a potential vehicle with zero velocity at the beginning of the nearest partly concealed patch as a safety measure.

Static road layouts, types of intersections and building positions are randomly generated. In addition, the urban traffic environment itself is dynamic and the number of traffic participants (vehicles and pedestrians), as well as their routes, are randomized. In every episode, there are between 30 and 120 simulated cars, as well as between 30 and 120 pedestrians, both chosen at random independently. We let each training episode run for 120 seconds or until a crash between the agent and another traffic participants prematurely occurs.
To facilitate reproducibility, we chose the DQN implementation from \textit{StableBaselines-3}~\cite{raffin2021sb}.

The agent selects its actions from the following discrete set of acceleration levels:
 $[-7, -3, -1.5, 0, 1.5, 3]$ ($\frac{m}{s^2})$. These values are based on realistic acceleration values, whereby the acceleration level $-7$ ($\frac{m}{s^2})$ can be interpreted as an \textit{emergency brake} action and can be selected by the shield if necessary.

As the agent reward function, we choose $r_{agent} := r_{collision} + \frac{1}{FPS}(r_{velocity} + r_{acceleration} + r_{intersection} +r_{shield} +r_{distance})$, where $FPS$ normalizes the sum of rewards to counterbalance any frame-dependencies. 

The collision penalty is independent of the frequency and is defined by 
\[
r_{collision} := -k_{c} \vert v \vert (\mathbbm{1}_{collision}\cup \mathbbm{1}_{near-collision}) - k_{c,abs},
\]
where $v$ stands for the agent's velocity, $\mathbbm{1}_{collision}$ and $\mathbbm{1}_{near-collision}$ are indicator functions that signal collisions (equally penalized), and $k_c, k_{c,abs} \in \mathbb{R}$ are hyperparameters that allow task-specific flexibility. Compared to~\cite{kurzer2021generalizing}, the speed at impact is also taken into account here, since we argue that collisions with high speed should be penalized harder than slow collisions. It should be noted that the current episode ends after one collision, which limits this penalty to at most once per episode.

We furthermore reward the agent for being near the allowed top speed ($v_{upper}$) with:
\begin{align*}
		r_{velocity} :=
		\begin{cases}
			-k_{v_{upper}} \vert v - v_{upper} \vert & \qquad \textrm{if } v > v_{upper} \\
			k_{v_{lower}} \vert v \vert & \qquad \textrm{else }
		\end{cases}
	\end{align*}
and accelerating proactively ($a$ is the current agent acceleration) with 
\[
r_{acceleration} := -k_a (2^{|a|}-1).
\]
Here, the intensity of acceleration is taken into account by punishing a reckless (potentially discomforting) driving style. The value of the scaling factor $k_a$ can be seen in table \ref{table:dqn-parameters}.
Since we don't want the agent to block an intersection at any time, we add a term 
\begin{align*}
    r_{intersection} := -k_{intersection},
    \end{align*}
that is a penalty that is only given if an agent vehicle is at or currently on an intersection.
Finally, we want to penalize the agent for using the shield and, instead, incentive proactive control of its velocity:
\begin{align*}
		r_{shield} := 
			-k_{shield} a_{shield} & \qquad \textrm{if } a_{shield}<a_{agent}
	\end{align*}
Here, $k_{shield} \in \mathbb{R}$ is a constant, $a_{shield}$ is the acceleration given by the shield, and $a_{agent}$ is the agent's choice of acceleration. This way, emergency brakes get penalized harder than other methods of deceleration.
To prevent the agent from keeping a huge safety distance and encouraging ``normal'' driving behavior, we lastly added a term
\[
		r_{distance} := k_{dist} \dfrac{d_{la}-d_{free}}{d_{la}} \qquad \textrm{if } a_{shield}\geq a_{agent} \wedge v > 0
  \]
 that gives a reward based on the distance to the preceding vehicle with $d_{la}$ being the distance the vehicle is able to see in front of it and $d_{free}$ being the length of unoccupied road.

The evaluation was done with 1000 episodes at $24FPS$. We evaluate the agent, called RL-Min, which was trained in an environment without buildings in an environment with buildings. Further training parameters can be seen in table~\ref{table:dqn-parameters}. In this context, all parameters $k_{}$ describe tunable hyperparameters.

\begin{table}[!htb]
    \centering
    \begin{tabular}{lrr}
        \toprule
        \multicolumn{2}{c}{\textbf{DQN Training Parameters}}\\ 
        %\cmidrule{1-2}
        Parameter & Value \\
        \midrule
        \ \\ \addlinespace
        Learning rate $\eta$ & 0.0001 \\ \addlinespace
        \textit{Memory-Replay} size & 500000  \\ \addlinespace
        \textit{Batch} size & 32 \\ \addlinespace
        Discount factor $\gamma$ & 0.99 \\ \addlinespace
        Exploration $\epsilon_{final}$ & 0.05 \\ \addlinespace
        Net shape & (112, 112) \\ \addlinespace
        Training steps & 25M \\
        \bottomrule
    \end{tabular}
    \begin{tabular}{lrr}
        \toprule
        \multicolumn{2}{c}{\textbf{Reward function Parameters}}\\ %\cmidrule{1-2}
        Parameter & Value \\
        \midrule
        $k_c$ & 3 \\ \addlinespace
        $k_{c,abs}$ & 25  \\ \addlinespace
        $k_{v_{upper}}$ & 0.06 \\ \addlinespace
        $k_{v_{lower}}$ & 0.03 \\ \addlinespace
        $k_a$ & 0.01 \\ \addlinespace
        $k_{intersection}$ & 0.2 \\ \addlinespace
        $k_{shield}$ & 0.1 \\ \addlinespace
        $k_{dist}$ & 0.2 \\
        \bottomrule
    \end{tabular}
    \caption{Parameters used in our experiments.}
    \label{table:dqn-parameters}
\end{table}

We compare our approach to several baselines: The TTC agent is an agent that behaves according to a rule set based on the \textit{time-to-collision} metric, similar to the Intelligent Driver Model (IDM), as proposed by Treiber et al. in~\cite{treiber2000idm}. The TTC-CREEP agent, which is a model that specializes on intersections with visibility restrictions, extends on the TTC agents with a modified behavior near intersections~\cite{isele2018navigating}.
The IER+-ACC is a new simple rule-based baseline developed in this work. The current acceleration of this agent follows a binary choice of accelerating with either $3 \frac{m}{s^2}$ if the velocity of the vehicle is below the threshold (\textit{here:} $50 km/h$) or $0 \frac{m}{s^2}$ if above.

The IER+-IDM agent is a novel \textit{baseline}. Similar to the IER+-ACC, it utilizes the shield of the RL agent's environment representation. However, its choice of acceleration is handled by the IDM. The IDM requires the distance to the next road user ahead to determine the acceleration. In this case, this distance can be derived directly from the IER+ state representation. The advantage of this is that the IDM also takes into account the imaginary hidden vehicles encoded in the state representation.

Three main aspects of particular importance to the evaluation were identified as part of this evaluation: Safety,
Speed and Consumption. The
consumption, however, was not considered in the work, although it should certainly play a major role due to environmental protection and economic efficiency.
Here, suitable evaluation metrics were chosen for all three aspects:
\begin{itemize}
    \item \textbf{Safety}: Ensuring safety is the most important aspect of autonomous driving. A low number of collisions indicate increased safety. At the same time, however, it must be noted that at an increased average speed, a greater distance is covered, and potentially more collisions can occur. For this reason, the ratio of collisions to average speed should also be considered. The smaller this ratio is, the safer the driving style of an agent.
    \item \textbf{Speed}: An agent should move at near maximum velocity in compliance with any safety requirements. The speed can be assessed directly from the recorded average speed of an agent.
    \item \textbf{Ecological/economical efficiency}: An autonomous vehicle should consume as little energy as possible. Instead of incorporating the energy consumption into our simulation, we instead compare the average positive acceleration in relation to the average speed, as this metric gives an indication of the efficiency in acceleration and thus in consumption. This means that if two agents have the same average speed and one agent accelerates significantly less, this agent tends to drive more predictively and efficiently. The smaller the ratio of acceleration to speed, the more efficient an agent is.
\end{itemize}

To test for generalizability, we train an agent (\textit{RL-Agent}) on large randomly generated traffic environments, such as those in figure \ref{fig:subprozesse}, and another agent (\textit{RL-Min-Agent}) on a minimal road setup with randomly generated concealing structures. We then evaluate the agents on a randomly chosen large road network, similar to figure \ref{fig:subprozesse}.

\begin{figure}[!h]
    \begin{center}
    \includegraphics[width=.9\columnwidth]{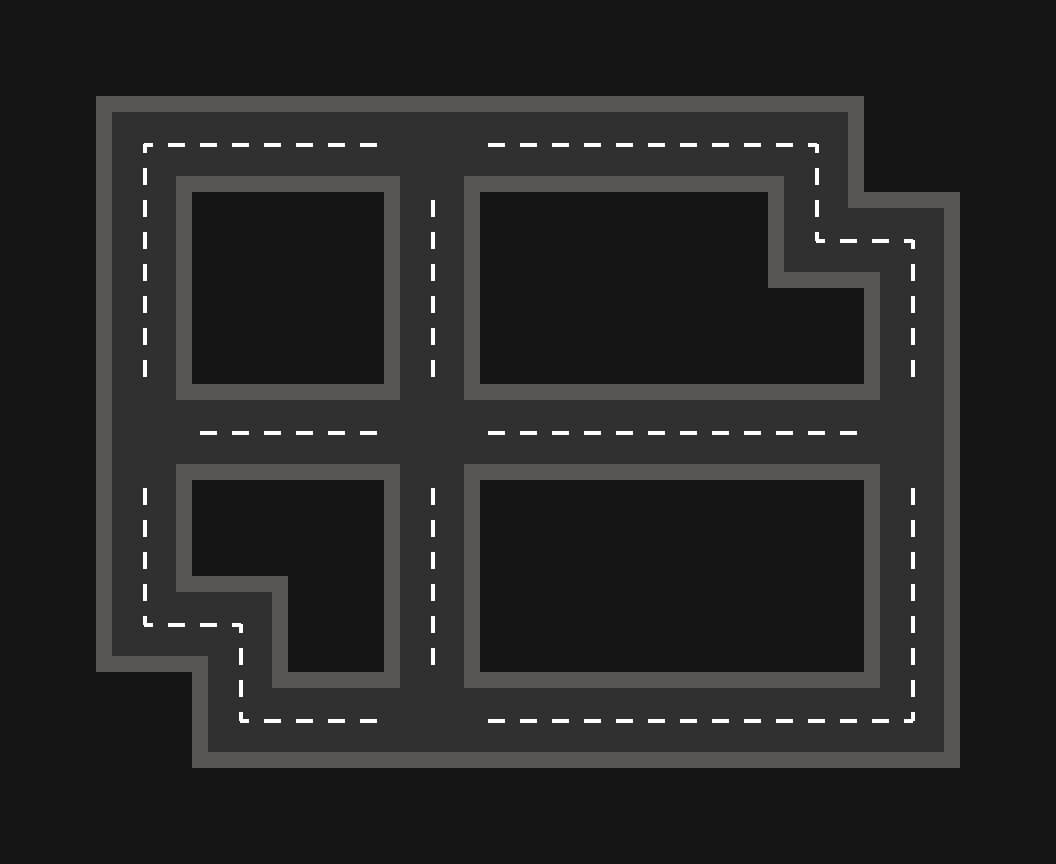}
    %\quelle{Aufnahme aus dem Simulator}
    \caption{Pre-defined minimal urban road network used for training \textit{RL-Min} in the \textit{PyAD-RL} simulator.}
    \label{fig:strassennetz_min_map}
    \end{center}
\end{figure}
The raw minimalistic network can be seen in figure \ref{fig:strassennetz_min_map}.
As can be seen in Table~\ref{table:results}, our approaches, \textit{IER+-IDM}, \textit{RL-Agent} and \textit{RL-Min}, outperform the baselines derived from the literature with a significantly lower collision rate. Even so, the two approaches themselves did not cause any collisions, but were only involved in a few collisions caused by simulated vehicles.
Our \textit{RL-Min} agent achieves a collision rate (collisions per average velocity)
 $61.5\%$ lower the \textit{IER+-ACC}, $42.5\%$ better than the \textit{IER+-IDM} baseline, $50.5\%$ better than the \textit{RL-Agent}, $97.7\%$ better than \textit{TTC-Creep} and $99\%$ better than \textit{TTC}.
We want to highlight here, that our novel baseline \textit{IER+-IDM }beats all previous rule-based baselines from literature significantly. Additionally, it operates at the highest energy efficiency, meaning the average acceleration per average vehicular velocity: $60\%$ better than \textit{IER+-ACC}, $12\%$ better compared to \textit{IER+-IDM}, $31.3\%$ better than \textit{TTC-Creep} and $8.3\%$ better than \textit{TTC}, only getting beaten by the \textit{RL-Agent} with a $22.2\%$ worse score, which can be explained by the different braking behavior due to different safety restrictions.

% Comment block
\iffalse
\begin{figure}[!htb]
    \centering
    \subfigure{
        \includegraphics[width=1\columnwidth]{media/results/094fd25c50303fc7d5549291fd3fae7247869801_col.car_velocity_agent.pdf}
    }
    %\subfigure{
    %    \includegraphics[width=1\columnwidth]{media/results/094fd25c50303fc7d5549291fd3fae7247869801_col.collisions.pdf}
    %}
    \subfigure{
        \includegraphics[width=1\columnwidth]{media/results/094fd25c50303fc7d5549291fd3fae7247869801_col.collision_efficiency_agent.pdf}
    }
   \subfigure{
        \includegraphics[width=1\columnwidth]{media/results/094fd25c50303fc7d5549291fd3fae7247869801_col.energy_efficiency_agent.pdf}
    }
     \caption{Comparison of the agent based on several metrics. Top: Average velocity (high is better). Middle: Collision rate (lower is better). Bottom: Energy consumption rate (lower is better).}
    \label{fig:eval2_velocity_rewards}
\end{figure}
\fi

\begin{table}[!htb]
\centering
\begin{tabular}{cccc}
\textbf{} & \textit{Avg. velocity} & \textit{Collision rate} & \textit{Energy eff. rate} \\
TTC       & \textbf{24.20}                  & 5.0                     & 0.024                     \\
TTC-Creep & 21.04                  & 2.43                    & 0.032                     \\
IER+-ACC  & 23.03                  & 0.13                    & 0.055                     \\
IER+-IDM  & 23.01                  & 0.09                    & 0.025                     \\
RL-Agent  & 19.80                  & 0.10                    & \textbf{0.018}                     \\
RL-Min   & 20.05                  & \textbf{0.05}                   & 0.022                    
\end{tabular}
\caption{Results from the generalization experiments.}
    \label{table:results}
\end{table}

All in all, we found that the implementation of the shield for the RL agent resulted in a significant improvement in safety, as measured by the collision rates. Additionally, it was shown that the state representation and the RL agent has a high generalization capability. This allowed for accelerated training on fixed minimal road networks and the transfer of the behavior to randomly generated large road networks.

%Comment this block if too much content
It must be noted that it would arguably take a lot of effort for the use of this state representation in reality. After all, this representation requires a correct prediction of the turning behavior of other road users as well as their speeds and distances. One possibility is the technology LiDAR~\cite{li2020lidar}, which detects the environment by emitting lasers. However, since LiDAR has issues with semantic recognition of objects, a combination of LiDAR and cameras would probably be necessary.

\section{Conclusion and Outlook}
In conclusion, our study has introduced a novel state space representation tailored for autonomous vehicles, enabling them to navigate urban environments with a strong emphasis on safety and compliance with traffic regulations. %Additionally, we have developed a specialized traffic simulator designed explicitly for rigorous testing of Reinforcement Learning (RL) algorithms.

Among our contributions (compare Section I), our research work comprises theoretical and practical aspects to achieve reproducibility and further development by the community:
\begin{enumerate}
\item \textbf{Effective Generalization:} Our research has demonstrated the capacity of our RL agent to generalize effectively. It excels baselines in unfamiliar, partially observable traffic scenarios.
\item \textbf{Customized Traffic Simulator:} We have created a dedicated traffic simulator, a valuable tool for assessing RL algorithms in complex, real-world-like scenarios. 
\end{enumerate}

Looking ahead, we envisage extending our work to encompass continuous action spaces, multi-agent scenarios, where agents collaborate through information sharing. This future direction holds the potential to enhance further the safety and efficiency of autonomous urban transportation systems, advancing the field of intelligent vehicles.
%We have proposed a state space representation that allows autonomous vehicles to navigate safely in urban environments while respecting traffic rules. We also presented a traffic simulator that was specifically developed to test RL algorithms. In our experiment, we have successfully shown that our RL agent generalizes well and is able to reach respectable average velocities in unknown and partially observable traffic scenarios.

\bibliographystyle{plain}
\bibliography{bibitems}
%\vspace{12pt}

\end{document}